\documentclass[a4paper,twocolumn,fleqn]{article}

\usepackage{ist}
\usepackage{graphicx}
\usepackage{amsmath}
\usepackage{amssymb}
\usepackage{cite}
\usepackage{placeins}              % \FloatBarrier to force float placement
\usepackage{hyperref}

\makeatletter
\long\def\@makecaption#1#2{%
  \ifx\@captype\@tablestring% table
    {\small\sffamily{\bfseries #1.}~~#2}\@tablecaptionsepspace
  \else % figure
    \@figurecaptionsepspace
    \setbox\@tempboxa\hbox{\scriptsize\sffamily\itshape {\bfseries #1.}~~ #2}%
    \ifdim \wd\@tempboxa >\hsize%
      \setbox\@tempboxa\hbox{\scriptsize\sffamily\itshape {\bfseries #1.}~~ }%
      \parbox[t]{\hsize}{\scriptsize\sffamily\itshape \noindent\unhbox\@tempboxa#2}%
    \else%
      \hbox to\hsize{\scriptsize\sffamily\itshape\box\@tempboxa\hfil}%
    \fi
  \fi}
\makeatother

\title{Beyond CCT: Multi-channel SPD recovery for per-shot camera color correction}

\author{
Shuwei Yue$^{1,*}$, Hongke Wei$^{2}$, and Qiao Chen$^{1}$\newline
$^{1}$Shenzhen Polytechnic University, Shenzhen, China\newline
$^{2}$Arashi Vision Inc. (Insta360), Shenzhen, China\newline
$^{*}$Corresponding author: shuwei.yue@connect.polyu.hk
}
\date{}
\begin{document}

\maketitle
\thispagestyle{empty}

%%%%%%%%%%%%%%%%%%%%%%%%%%%%%%%%%%
% Abstract
%%%%%%%%%%%%%%%%%%%%%%%%%%%%%%%%%%

\begin{abstract}
Modern cameras correct color via two or three pre-calibrated $3{\times}3$ matrices interpolated by Correlated Color Temperature (CCT). This collapses the illuminant SPD to a scalar. LEDs, fluorescents, and mixed sources sit off the Planckian locus, and metameric illuminants with the same chromaticity demand different optimal CCMs that a CCT pipeline cannot distinguish. We recover the illuminant SPD instead --- non-negative least squares against a basis dictionary, conditioned on the camera white-point and a multi-channel ambient-light sensor (ALS). A closed-form $3{\times}3$ CCM is then fit per shot. On a synthetic benchmark of 300 illuminants $\times\,3{,}100$ reflectances, mean angular error drops from $1.89^{\circ}$ to $1.20^{\circ}$, with a $1.7\times$ tail reduction on the LED-like off-Planckian group ($n{=}45$) where metameric ambiguity is tangible. The gain holds under realistic ALS imperfections and observation noise, and the same off-Planckian advantage carries over to $1{,}684$ real measured SPDs.
\end{abstract}

%%%%%%%%%%%%%%%%%%%%%%%%%%%%%%%%%%
\section{Introduction}
%%%%%%%%%%%%%%%%%%%%%%%%%%%%%%%%%%

The Color Correction Matrix (CCM) is a $3{\times}3$ linear map from white-balanced camera RGB to a colorimetric standard (CIE XYZ or sRGB). The optimal CCM depends on the scene illuminant. The standard Adobe DNG pipeline~\cite{dng_spec} stores two reference CCMs, one at $T_1{=}2856$~K and one at $T_2{=}6500$~K. At runtime it interpolates the two by the estimated Correlated Color Temperature (CCT). The illuminant SPD $E(\lambda)$ --- a function over wavelength --- is collapsed to a single scalar.

\noindent This works on Planckian illuminants. Blackbody radiators are fully described by temperature, so CCT and the SPD are equivalent. Modern indoor and architectural lighting often sits off this locus. LEDs, fluorescents, and mixed sources do not follow the blackbody curve. Two sources can share $xy$ chromaticity to within $\Delta_{xy}{<}0.005$ yet differ in normalized SPD shape by more than $30^{\circ}$. Both drive the same white-point reading and the same CCT estimate. They still need different optimal CCMs. CCT-based interpolation cannot tell them apart.

\noindent We present a spectral-aware per-shot CCM that replaces the CCT scalar with a recovered SPD. We recover the SPD from two cheap observations: the camera's own white-point response and a multi-channel ambient-light sensor (ALS). The ALS is a small commodity chip with 8 to 14 narrow-band photodiodes, already used in mobile phones for ambient sensing. With $K{+}3$ observations against a small SPD basis, recovery is a non-negative least squares problem. The recovered SPD then synthesizes a virtual color chart, and we fit a scene-specific CCM to that chart in closed form.

\begin{figure}[!t]
\vspace{-2mm}
\centering
\includegraphics[width=0.95\linewidth]{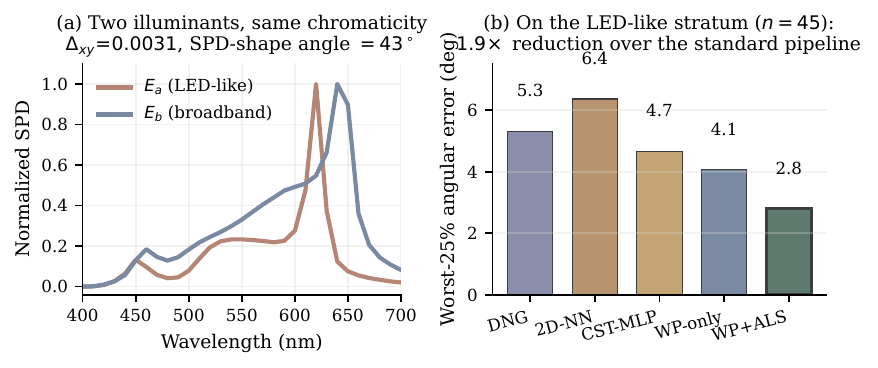}
\caption{The metameric problem and our fix. (a) Two real EMPIR illuminants share $xy$ chromaticity to within $\Delta_{xy}{=}0.003$ but differ in normalized SPD shape by $43^{\circ}$. A CCT-driven CCM cannot tell them apart, yet they need different optimal CCMs. (b) On the LED-like Duv group of our benchmark ($n{=}45$, statistically meaningful), our spectral pipeline cuts worst-25\% angular error from $5.3^{\circ}$ (Adobe DNG) to $\mathbf{2.8^{\circ}}$ --- a $1.9{\times}$ reduction. A more extreme group ($|\mathrm{Duv}|{\geq}0.03$, $n{=}5$) shows a larger directional gain that we report only for context.}
\label{fig:teaser}
\vspace{-3mm}
\end{figure}

\noindent Figure~\ref{fig:teaser} previews the argument: a real metameric pair on the left, the worst-case error reduction on the right. The full method comparison and ablations are in Sec.~4; a real-data sanity check on the WP-only fallback (without ALS hardware) is in Sec.~4.5.

\noindent \textbf{Contributions}~The non-parametric NNLS form lifts CCM conditioning from chromaticity to $K{+}3$ dimensions. Against six baselines on 300 EMPIR illuminants $\times\,3{,}100$ CAVE reflectances, the spectral advantage is most reliable on the LED-like off-Planckian group ($n{=}45$), where worst-25\% error drops by $1.7{\times}$ ($4.79^{\circ}\!\to\!2.80^{\circ}$) relative to the strongest non-spectral baseline (CST-MLP$_{3\times6}$, the recent $xy$-MLP of Tedla \emph{et al.}~\cite{tedla2025}); a smaller extreme subset ($n{=}5$) shows a larger directional gain. A channel-count ablation finds a sharp knee at $K{=}8$, matching commercially available multispectral sensors~\cite{ams_as7341}. A metameric-pair stress test isolates SPD-recovery accuracy from downstream CCM behavior.

%%%%%%%%%%%%%%%%%%%%%%%%%%%%%%%%%%
\section{Related work}
%%%%%%%%%%%%%%%%%%%%%%%%%%%%%%%%%%

Three lines of work bear on a spectral-aware CCM. Classical color constancy estimates an illuminant chromaticity but stops short of the CCM stage. CCT- and $xy$-conditioned CCM pipelines scalarize the illuminant to one or two numbers, which a metameric pair defeats by construction. Spectral-recovery work tackles reflectance rather than the illuminant or the CCM. We position our method against each in turn.

\noindent \textbf{Classical color constancy}~Statistical priors --- gray-world~\cite{buchsbaum_grayworld}, gray-edge~\cite{vandeweijer_grayedge} --- and chromaticity-prior approaches that exploit saturated colors~\cite{yue2023purecolor} estimate an illuminant chromaticity but stop there; downstream CCM remains a separate, illuminant-conditioned step.

\noindent \textbf{CCT-based CCM}~The DNG two-CCT pipeline~\cite{dng_spec} and Karaimer \& Brown's three-CCT extension~\cite{karaimer_brown_2018} both parameterize the illuminant in 1D for the CCM stage. We treat upstream AWB as orthogonal to the CCM-stage spectral conditioning we address here.

\noindent \textbf{Beyond CCT}~Tedla \emph{et al.}~\cite{tedla2025} share our diagnosis that off-Planckian sources need richer conditioning than CCT, and use $xy$ chromaticity (2D) as input to a small MLP that predicts a $3{\times}3$ or $3{\times}6$ root-polynomial~\cite{finlayson_rootpoly} mapping. Their input is still the white point projected to chromaticity, so two SPDs sharing $xy$ (a metameric pair) yield identical CCMs by construction. We lift the conditioning dimension from 2 to $K{+}3$ where $K$ is the ALS channel count: a multi-channel sensor resolves the spectral shape that the $xy$ projection discards.

\noindent \textbf{Spectral upsampling and HSI}~Recovering reflectance spectra from RGB~\cite{arad_ben_shahar} and learning illuminant SPDs from hyperspectral data~\cite{kaust_hsi} solve neighboring problems; we recover the illuminant SPD given known camera response and an auxiliary multi-channel observation.

%%%%%%%%%%%%%%%%%%%%%%%%%%%%%%%%%%
\section{Method}
%%%%%%%%%%%%%%%%%%%%%%%%%%%%%%%%%%

We treat per-shot CCM design as a spectral-recovery problem. Each shot follows three stages. First, we recover an illuminant SPD $\hat E$ from the camera white point and a multi-channel ALS reading by a small non-negative least squares against a fixed SPD basis. Second, we plug $\hat E$ into the imaging equations to synthesize a virtual 24-patch chart in raw camera RGB and reference CIE XYZ. Third, we fit a $3{\times}3$ CCM to that chart in closed-form least squares, apply it to the captured pixels, and chromatically adapt to a D65 reference frame. The rest of this section formalizes each stage and the ALS filter bank.

\noindent \textbf{Forward model}~We discretize wavelength on $[400,700]$~nm at $10$~nm ($N_\lambda{=}31$). Let $E\in\mathbb{R}^{N_\lambda}$ be the unknown SPD; let $\mathbf{S}\in\mathbb{R}^{N_\lambda\times 3}$ be the camera SSF, $\mathbf{C}\in\mathbb{R}^{N_\lambda\times 3}$ the CIE 1931 CMFs, $\mathbf{A}\in\mathbb{R}^{N_\lambda\times K}$ the ALS filter bank, and $\mathbf{R}\in\mathbb{R}^{N_\lambda\times 24}$ a fixed 24-patch ColorChecker-style reflectance set. The camera white-point and ALS observations are wavelength integrals of $E$ against the SSF and ALS filters, respectively. Per-patch camera RGB and CIE XYZ are the same integrals weighted by the patch reflectance.

\noindent \textbf{NNLS recovery over an SPD basis}~We constrain $E$ to the non-negative span of a basis $\mathbf{B}\in\mathbb{R}^{N_\lambda\times M}$. To build $\mathbf{B}$, we max-normalize each EMPIR PhotoLED SPD~\cite{empir_database} so basis atoms describe shape, not magnitude; absolute scale is absorbed into the recovered coefficients $\hat{\boldsymbol\alpha}$, and we observe condition numbers $\kappa(\mathbf{F})\!<\!50$ across the test illuminant set. We pick $M{=}30$ representative atoms by $k$-means on a held-out 200-illuminant pool for the main experiments; Table~\ref{tab:robust}(a) shows results are insensitive to $M\!\in\![10,100]$. Stacking the camera white-point and ALS readings into $\mathbf{y}{=}[\mathrm{als};\mathrm{wp}]^\top\!\in\!\mathbb{R}^{K{+}3}$ and the corresponding forward operator $\mathbf{F}{=}[\mathbf{A}^\top\mathbf{B};\mathbf{S}^\top\mathbf{B}]$, we solve
\begin{equation}
\hat{\boldsymbol\alpha} = \arg\min_{\boldsymbol\alpha\geq 0}\,\lVert\mathbf{F}\boldsymbol\alpha-\mathbf{y}\rVert_2^2,\quad \hat E=\mathbf{B}\hat{\boldsymbol\alpha}.
\label{eq:nnls}
\end{equation}
We use the active-set NNLS in \texttt{scipy.optimize.nnls}. Non-negativity alone regularizes the inversion; we add no Tikhonov term. Train and test illuminant sets are disjoint and grouped by $|\mathrm{Duv}|$ (Sec.~4.1).

\noindent \textbf{Per-shot virtual calibration}~With $\hat E$ in hand, we synthesize a virtual 24-patch chart by plugging $\hat E$ into the forward model. For each patch $k$, we obtain a synthesized camera RGB $\hat{\mathbf{r}}_k$ and CIE XYZ $\hat{\mathbf{x}}_k$ as wavelength integrals of $\hat E\cdot\mathbf{R}_k$ against $\mathbf{S}$ and $\mathbf{C}$ respectively. We fit $\mathbf{M}\in\mathbb{R}^{3\times 3}$ as an XYZ$\to$RGB CCM by closed-form least squares. This direction matches the DNG \texttt{ColorMatrix} convention. The camera RGB and XYZ are defined up to a free exposure / normalization scale that propagates into $\mathbf{M}$; we pin the green-row middle entry $M_{22}{=}1$ for gauge-fixing, one of several admissible normalizations:
\begin{equation}
\mathbf{M}=\arg\min_{\mathbf{M}}\sum_{k=1}^{24}\lVert\hat{\mathbf{r}}_k-\mathbf{M}\,\hat{\mathbf{x}}_k\rVert_2^2\quad\text{s.t.}\;M_{22}{=}1.
\label{eq:lsq}
\end{equation}
At inference, we map a captured RGB $\mathbf{r}$ to scene-frame XYZ by $\mathbf{x}(\mathbf{r}){=}\mathbf{M}^{-1}\mathbf{r}$. The matrices we recover have condition numbers in the $10^1$--$10^2$ range, so the inversion is numerically stable. We then apply Bradford chromatic adaptation~\cite{bradford_cat} from $\hat E$'s white point $\mathbf{C}^\top \hat E$ to D65 for evaluation.

\begin{figure}[!t]
\vspace{-2mm}
\centering
\includegraphics[width=0.95\linewidth]{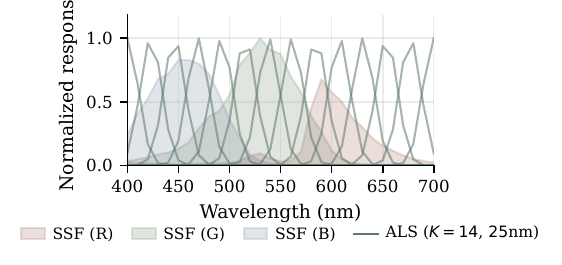}
\caption{Nikon D40 spectral sensitivity (filled bands) overlaid with the $14$-channel ALS filter bank used in our experiments.}
\label{fig:filterbank}
\vspace{-3mm}
\end{figure}

\noindent \textbf{ALS filter bank}~Our main experiments use $K$ idealized Gaussian filters with FWHM 25~nm and peaks uniformly spaced over 400--700~nm (Fig.~\ref{fig:filterbank}); this serves as a clean control for the channel-count ablation. An AS7341-like response with asymmetric peaks at the AS7341 channel centers, $5\%$ adjacent-channel crosstalk, and $3\%$ DC offset is reported in Sec.~4.3.

%%%%%%%%%%%%%%%%%%%%%%%%%%%%%%%%%%
\section{Experiments}
%%%%%%%%%%%%%%%%%%%%%%%%%%%%%%%%%%

We evaluate the pipeline on a synthetic benchmark of 300 illuminants $\times$ 3{,}100 reflectances against six baselines (CCT-driven, $xy$-conditioned, spectral), then probe robustness, OOD generalization, a metameric-pair stress test, and finally a measured-SPD evaluation plus real-RAW sanity check.

\begin{figure}[!t]
\vspace{-2mm}
\centering
\includegraphics[width=0.95\linewidth]{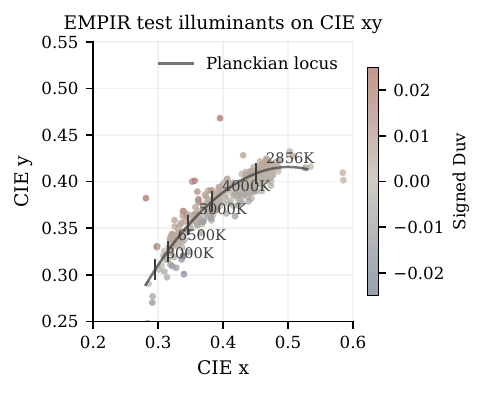}
\caption{EMPIR test illuminants on CIE 1931 $xy$, color-coded by signed Duv. The dataset spans Planckian sources, daylight phases, and significantly off-locus LEDs.}
\label{fig:xy}
\vspace{-3mm}
\end{figure}

The camera spectral sensitivity is the Nikon D40 from~\cite{jiang2013camspec}. Test illuminants are 300 EMPIR sources disjoint from the basis pool (random seed $0$), split by signed Duv into Planckian ($|\mathrm{Duv}|{<}0.003$, $n{=}156$), mild ($0.003{-}0.01$, $n{=}94$), LED-like ($0.01{-}0.03$, $n{=}45$), and extreme ($\geq 0.03$, $n{=}5$; see Fig.~\ref{fig:xy}). Test reflectances are 100 random pixels per scene over 31 CAVE scenes~\cite{cave_database}, $3{,}100$ vectors total. The calibration target is a 24-patch ColorChecker-style reflectance set distinct from the CAVE test reflectances, so the evaluation is a cross-distribution test. We report per-pair angular error on D65 XYZ as both the mean and the worst-25\% mean (average of the upper quartile, which exposes the metameric tail); CIE $\Delta E_{2000}$~\cite{deltaE2000} tracks the same ranking and is reported inline.

\noindent \textbf{Baselines}~\emph{Oracle}: per-illuminant least-squares CCM with the true SPD known; a non-causal upper bound. \emph{2-CST DNG}~\cite{dng_spec}: Adobe ColorMatrix1/2 with $1/T$ interpolation by McCamy CCT~\cite{mccamy_cct}. \emph{2-CST SELF}: same interpolation with self-calibrated matrices (D40 SSF $\times$ Planckian $\times$ ColorChecker reflectance). \emph{2D-NN}: nearest-neighbor lookup in $xy$ over basis-illuminant CCMs. \emph{CST-MLP}~\cite{tedla2025}: a reimplementation of both heads of Tedla \emph{et al.}: a 2-layer 32-unit MLP from $xy$ to either (i) the 8 free entries of a $3{\times}3$ CCM (``CST-MLP$_{3\times3}$''), or (ii) an $18$-element $3{\times}6$ matrix that maps Finlayson root-polynomial features~\cite{finlayson_rootpoly} to XYZ (``CST-MLP$_{3\times6}$''). Both heads trained on the 200 basis illuminants for $2{,}000$ Adam steps at $\mathrm{lr}{=}10^{-3}$ with per-patch cosine loss. \emph{Spectral WP-only}: our framework with $K{=}0$. \emph{Spectral WP$+$ALS}: the full method with $K{=}14$.

\noindent \textbf{Main result: synthetic benchmark by Duv}~Table~\ref{tab:main} reports mean and worst-25\% angular error per Duv group. WP$+$ALS is the best non-Oracle method at every group. Against the strongest non-spectral baseline (CST-MLP$_{3\times6}$, $1.89^{\circ}$ overall), WP$+$ALS reduces mean error by $36\%$ ($1.20^{\circ}$). The headline gain is statistically grounded on the LED-like group ($n{=}45$), where the metameric problem is tangible: worst-25\% drops from $4.79^{\circ}$ to $2.80^{\circ}$, a $1.71{\times}$ reduction. The Extreme group's tail is even larger but rests on only $n{=}5$ illuminants and is reported directionally. The $xy$-conditioned baselines (2D-NN, CST-MLP) beat the CCT-driven baselines at low Duv, confirming Tedla \emph{et al.}'s diagnosis. They fall apart off the locus, where WP$+$ALS has $K{+}3$ conditioning dimensions instead of $2$. The 3$\times$6 root-polynomial extension of Tedla shrinks but does not close the gap (1.89$^\circ$ vs.\ 1.20$^\circ$ overall, 2.27$^\circ$ vs.\ 1.31$^\circ$ on LED-like). On Planckian, ALS adds only $0.50^{\circ}$ over WP-only --- positive but small.

\begin{table*}[h!]
\centering
\caption{Mean and worst-25\% angular error (degrees) per Duv group on $300{\times}3{,}100$ (illuminant, reflectance) pairs after Bradford CAT to D65. $3{\times}3$ and $3{\times}6$ are the linear and root-polynomial heads of Tedla \emph{et al.}'s CST-MLP~\cite{tedla2025} ($3{\times}6$ uses Finlayson root-polynomial features~\cite{finlayson_rootpoly}). Best non-Oracle entry per row in \textbf{bold}. The Extreme group has only $n{=}5$ illuminants; its worst-25\% averages 1--2 samples and is reported for completeness, not statistical inference. Statistical weight rests on the LED-like group ($n{=}45$). Paired bootstrap on the WP$+$ALS vs CST-MLP$_{3\times6}$ mean gap (illuminant-level resampling, $B{=}1000$): overall $\Delta\!=\!+0.70^{\circ}$, 95\% CI $[+0.40,\,+1.19]^{\circ}$; LED-like $\Delta\!=\!+0.97^{\circ}$, 95\% CI $[+0.67,\,+1.37]^{\circ}$; Extreme $\Delta\!=\!+5.62^{\circ}$, 95\% CI $[+0.01,\,+14.56]^{\circ}$ (the wide CI is the $n{=}5$ caveat made explicit).}
\vspace{3mm}
\label{tab:main}
\resizebox{\textwidth}{!}{
\begin{tabular}{l|ccc|ccc|cc}
\hline
 & \multicolumn{3}{c|}{\textbf{CCT-driven}} & \multicolumn{3}{c|}{\textbf{$xy$-conditioned}} & \multicolumn{2}{c}{\textbf{Spectral (ours)}} \\
$|\mathrm{Duv}|$ group
 & Oracle & DNG~\cite{dng_spec} & SELF
 & 2D-NN & $3{\times}3$~\cite{tedla2025} & $3{\times}6$
 & WP-only & \textbf{WP$+$ALS} \\
\hline
\multicolumn{9}{l}{\textit{Mean angular error}} \\
Planckian $<\!0.003$ ($n{=}156$) & 0.95 & 2.45 & 3.18 & 1.75 & 1.79 & 1.33 & 1.55 & \textbf{1.05} \\
Mild $0.003\!-\!0.01$ ($n{=}94$)  & 0.87 & 2.73 & 3.44 & 2.01 & 1.99 & 2.21 & 1.82 & \textbf{1.21} \\
LED-like $0.01\!-\!0.03$ ($n{=}45$) & 0.81 & 3.09 & 4.26 & 2.78 & 2.41 & 2.27 & 2.18 & \textbf{1.31} \\
Extreme $\geq\!0.03$ ($n{=}5$)   & 2.36 & 11.08 & 26.75 & 6.60 & 14.31 & 10.11 & 9.23 & \textbf{4.49} \\
\textbf{Overall} ($n{=}300$)     & 0.93 & 2.77 & 3.81 & 2.07 & 2.15 & 1.89 & 1.86 & \textbf{1.20} \\
\hline
\multicolumn{9}{l}{\textit{Worst-25\% angular error (mean of upper quartile)}} \\
Planckian $<\!0.003$           & 2.02 & 4.29 & 5.33 & 3.74 & 3.61 & 2.77 & 3.21 & \textbf{2.27} \\
Mild $0.003\!-\!0.01$          & 1.92 & 5.05 & 6.47 & 4.57 & 4.15 & 5.97 & 3.92 & \textbf{2.83} \\
LED-like $0.01\!-\!0.03$       & 1.88 & 5.32 & 8.50 & 6.37 & 4.67 & 4.79 & 4.08 & \textbf{2.80} \\
Extreme $\geq\!0.03$            & 7.10 & 35.88 & 88.34 & 18.86 & 45.66 & 33.39 & 30.97 & \textbf{12.63} \\
\textbf{Overall}                & 2.06 & 5.29 & 7.73 & 4.71 & 4.74 & 4.68 & 4.08 & \textbf{2.74} \\
\hline
\end{tabular}
}
\vspace{3mm}
\end{table*}

\noindent The visual gap is pronounced on saturated chart patches: metameric ambiguity translates most directly into hue and saturation errors, where WP$+$ALS approaches the Oracle floor while WP-only and the CCT baselines retain large per-patch residuals on red, orange, and blue primaries (visible in the worst-25\% column of Table~\ref{tab:main}, dominated by these patches).

\noindent \textbf{Perceptual color difference}~Mean $\Delta E_{2000}$~\cite{deltaE2000} for DNG / CST-MLP$_{3\times3}$ / WP-only / WP$+$ALS: overall $2.10/1.55/1.12/\mathbf{0.69}$; LED-like ($n{=}45$) $2.20/1.65/1.32/\mathbf{0.80}$; Extreme ($n{=}5$, directional) $5.94/8.66/5.66/\mathbf{2.78}$. Same ranking as angular error.

\noindent \textbf{Robustness and basis representativeness}~We test the proposed pipeline along four axes (Table~\ref{tab:robust}). \emph{(a) Channel count $\times$ basis size}: sweeping $K\!\in\!\{0,4,8,14,28\}$ against $M\!\in\!\{10,30,50,100\}$ on the off-Planckian subset, the $K{=}8$ knee is robust across $M$ (within $0.16^\circ$ of the row minimum at every $M$), consistent with the classical 5--8-dimensional rank of natural illuminant SPDs. \emph{(b) Realistic ALS filter shape}: an AS7341-like response --- asymmetric peaks at the AS7341 channel centers, $5\%$ crosstalk, $3\%$ DC offset; a conservative non-idealization, not manufacturer-measured curves --- at $K{=}8$ raises mean error from $1.21^{\circ}$ to $1.32^{\circ}$ ($+8.7\%$). \emph{(c) Observation noise}: multiplicative Gaussian noise on $(\mathrm{wp}, \mathrm{als})$ at $\sigma\!\in\!\{1,3,5\}\%$ degrades WP$+$ALS mean error from $1.20^\circ$ to $1.23^\circ$ ($+2.8\%$ at $5\%$); the WP$+$ALS--DNG crossover would require $\sigma\!\approx\!50\%$. \emph{(d) Cross-database (out-of-distribution)}: on $23$ CIE reference illuminants (FL$_{1\text{--}9}$, HP$_{1\text{--}5}$, LED-B/V/RGB) held out from EMPIR, an EMPIR-only basis under-performs DNG ($2.75^\circ$ vs.\ $2.21^\circ$): the basis cannot represent fluorescent Hg lines. Augmenting the pool with $7$ CIE references (one per major class) and re-running $k$-means restores the spectral lead in this controlled test ($2.00^\circ$ vs.\ DNG $2.25^\circ$ on the remaining $16$ hold-outs). A small class-representative augmentation can substantially improve OOD behavior; the count and selection strategy are tuned for this benchmark.

\begin{table}[h!]
\centering
\caption{Robustness of WP$+$ALS along four axes. Numbers are mean / worst-25\% angular error in degrees on the relevant test set; for reference, DNG is $2.77/5.29$ on the same scale (Table~\ref{tab:main}). Cross-DB OOD test set: $23$ hold-out CIE reference illuminants (FL/HP/LED), with the bottom row using a basis pool augmented by $7$ CIE reference SPDs.}
\vspace{3mm}
\label{tab:robust}
\begin{tabular}{lc}
\hline
\textbf{Axis / condition} & \textbf{mean / worst-25 ($^{\circ}$)} \\
\hline
Baseline (ideal $K{=}14$, no noise)   & $1.20\,/\,2.74$ \\
AS7341-like filters ($K{=}8$)         & $1.32\,/\,3.25$ \\
$5\%$ observation noise               & $1.23\,/\,2.83$ \\
OOD, EMPIR-only basis                 & $2.75\,/\,5.40$ \\
OOD, basis $+\,7$ CIE refs            & $\mathbf{2.00\,/\,4.21}$ \\
\hline
\end{tabular}
\vspace{3mm}
\end{table}

\noindent \textbf{Metameric pair stress test}~We pick 50 EMPIR pairs $(E_a, E_b)$ with $\Delta_{xy}{<}0.005$ but SPD-shape angle larger than $30^{\circ}$ (where SPD-shape angle is $\arccos(\langle E_a,E_b\rangle/\lVert E_a\rVert\lVert E_b\rVert)$ in degrees on the unit-normalized SPDs). Both sides are held out from the basis. WP-only NNLS \emph{collapses} the pair (recovered pair angle $12.6^{\circ}$, true $35^{\circ}$); WP$+$ALS recovers $25.2^{\circ}$, with per-side SPD error $9.3^{\circ}$ vs.\ $24.6^{\circ}$ --- a $2.6{\times}$ improvement. Fig.~\ref{fig:teaser}(a) shows a comparable pair. On these cherry-picked pairs the per-patch CCM error is slightly higher for WP$+$ALS ($1.87^{\circ}$) than WP-only ($1.29^{\circ}$): WP-only converges on a single ``compromise'' SPD that fits both sides forgivingly, while WP$+$ALS exposes basis-discretization error. Better SPD recovery does not linearly translate into lower CCM error; the basis and CCM objective are not perfectly aligned. What WP$+$ALS uniquely provides is the ability to identify the correct side of a metamer; on the broader benchmark above, where reflectances vary freely, this converts to a measurable CCM-error gain.

\begin{figure*}[!t]
\vspace{-2mm}
\centering
\includegraphics[width=0.92\textwidth]{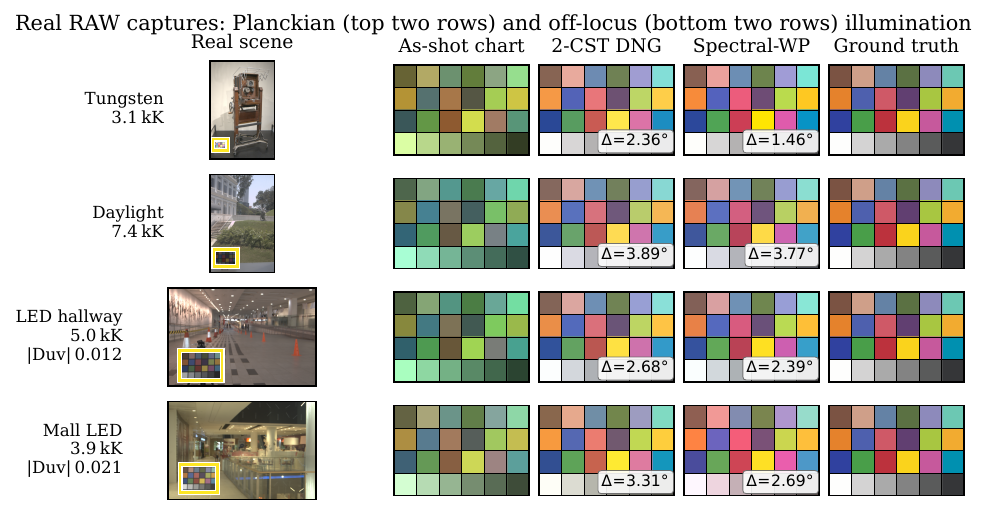}
\caption{Four real Nikon D40 captures. Top two rows near-Planckian (tungsten museum, daylight outdoor). Bottom two rows off the locus (LED hallway $|\mathrm{Duv}|{=}0.012$, mall LED $|\mathrm{Duv}|{=}0.021$). Left column shows the scene with the chart highlighted. The four right columns show the 24-patch chart \emph{as-shot} (raw camera response, before any white-balance, so each row shows its own illuminant cast), corrected by 2-CST DNG, by our spectral white-point pipeline, and the D65 ground truth. Per-shot mean angular error $\Delta$ relative to the D65 GT is annotated on the two corrected columns; the as-shot column is shown for context only. Spectral-WP edges out or matches DNG, with the largest gain on the most distant illuminant (tungsten, $1.46^{\circ}$ vs.\ $2.36^{\circ}$).}
\label{fig:realdata}
\vspace{-3mm}
\end{figure*}

\noindent \textbf{Measured-SPD evaluation and real RAW sanity check}~We replace the EMPIR test illuminants with the $1{,}684$ spectroradiometer-measured SPDs from Beyond RGB~\cite{beyondrgb} (1{,}209 lab + 475 outdoor); the rest of the protocol (Nikon~D40 SSF, CAVE reflectance, AS7341-like ALS synthesized from the measured SPD) is unchanged. This validates the illuminant side of the pipeline, not the full RGB$+$ALS capture loop. The same per-Duv pattern survives the swap to measured SPDs in both mean and worst-25\% (Table~\ref{tab:beyondrgb}): WP$+$ALS leads on LED-like and Extreme groups in both metrics, while CST-MLP$_{3\times6}$ wins on Planckian and Mild. The LED-like advantage on Beyond RGB is smaller than on the synthetic benchmark ($0.24^{\circ}$ mean / $0.17^{\circ}$ worst-25\% here vs $0.96^{\circ}$ / $1.99^{\circ}$ in Table~\ref{tab:main}); we attribute this to the broader test-illuminant distribution and the sub-optimality of the LED-biased EMPIR basis on daylight-dominated outdoor scenes. Spectral conditioning pays off where chromaticity is ambiguous, but the size of that pay-off depends on how off-locus the test set is.

\begin{table}[h!]
\centering
\caption{Beyond RGB~\cite{beyondrgb} mean and worst-25\% angular error (degrees) by Duv group on $1{,}684$ measured-SPD scenes. Groups: Plk = Planckian ($n{=}868$), Mild ($n{=}473$), LED = LED-like ($n{=}276$), Ext = Extreme ($n{=}67$). Best per row in \textbf{bold}. WP$+$ALS leads on LED and Ext in both metrics; the worst-25\% advantage on LED is smaller than the mean advantage ($0.17^{\circ}$ vs $0.24^{\circ}$ over CST-MLP$_{3\times6}$), reflecting the broader-distribution test illuminants compared to the synthetic benchmark.}
\vspace{3mm}
\label{tab:beyondrgb}
\begin{tabular}{lcccc}
\hline
\textbf{Method} & \textbf{Plk} & \textbf{Mild} & \textbf{LED} & \textbf{Ext} \\
\hline
\multicolumn{5}{l}{\textit{Mean angular error}} \\
DNG                  & 2.12 & 1.91 & 2.51 & 6.97 \\
CST-MLP$_{3\times6}$ & \textbf{1.72} & \textbf{1.67} & 2.13 & 9.50 \\
WP-only              & 2.67 & 2.31 & 3.16 & 5.76 \\
WP$+$ALS             & 2.17 & 1.83 & \textbf{1.89} & \textbf{5.14} \\
\hline
\multicolumn{5}{l}{\textit{Worst-25\% angular error}} \\
DNG                  & 4.40 & 3.71 & 5.40 & 21.61 \\
CST-MLP$_{3\times6}$ & \textbf{3.66} & \textbf{3.67} & 4.50 & 31.64 \\
WP-only              & 4.84 & 4.69 & 6.26 & 15.70 \\
WP$+$ALS             & 4.56 & 4.01 & \textbf{4.33} & \textbf{15.50} \\
\hline
\end{tabular}
\vspace{3mm}
\end{table}

\noindent We sanity-check on 117 real Nikon D40 captures with per-image white-point measurements but no ALS hardware (Fig.~\ref{fig:realdata} shows four). The dataset is $84\%$ near-Planckian; all three WP-only methods land within $0.5^{\circ}$ in mean angular error (2-CST SELF $3.00^{\circ}$, Spectral WP-only $3.09^{\circ}$, 2-CST DNG $3.42^{\circ}$). This is the predicted behavior on a near-Planckian distribution: the WP-only fallback recovers the SPD's 1D position on the locus, and the ALS adds nothing it can resolve there. End-to-end validation of WP$+$ALS on synchronized RGB$+$ALS captures remains future work.

%%%%%%%%%%%%%%%%%%%%%%%%%%%%%%%%%%
\section{Discussion}
%%%%%%%%%%%%%%%%%%%%%%%%%%%%%%%%%%

We close with one implication from the channel-count ablation, then the limitations.

\noindent \textbf{Implication: where 8 channels come from}~The knee of the WP$+$ALS error curve sits at $K{=}8$, robust across basis sizes $M\!\in\![10,100]$ (Sec.~4.3). This matches the classical 5--8-dimensional rank of natural illuminant SPDs and the channel count of commodity multispectral ALS chips~\cite{ams_as7341}: spectral information past 8 narrow bands gives diminishing returns on the CCM endpoint. A practical reading: an 8-channel commodity ALS provides enough spectral conditioning for the off-Planckian band where $xy$-conditioned baselines fail.

\noindent \textbf{When does better SPD recovery yield a better CCM?}~The metameric stress test (Sec.~4.5) shows an apparent paradox: WP$+$ALS recovers metameric SPDs $2.6\times$ more accurately than WP-only, yet per-patch CCM error on those cherry-picked pairs is slightly worse ($1.87^{\circ}$ vs $1.29^{\circ}$). The mechanism is that WP-only converges on a single ``compromise'' SPD that fits both sides of a metameric pair forgivingly; WP$+$ALS resolves the pair but then pays basis-discretization error on a single side. SPD-recovery accuracy is therefore \emph{not sufficient} for CCM accuracy. It is, however, \emph{necessary} on the main benchmark, where reflectances span a wide distribution: there, the ability to identify the correct metamer side converts directly into the $1.7\times$ worst-25\% reduction on LED-like illuminants reported in Sec.~4.1. The stress test is a pathological boundary; the main benchmark is the deployment proxy.

\noindent \textbf{Limitations and future work}~We isolate the spectral-conditioning contribution by holding the prediction head fixed at the non-parametric NNLS form throughout. The orthogonal axis --- training a parametric head (MLP, root-polynomial, etc.) on the same $K{+}3$-dimensional input --- is the learned-amortization extension noted as future work; the CST-MLP comparators receive only $xy$ chromaticity by their original design~\cite{tedla2025}. The ALS observations are simulated; the AS7341-like and noise analyses cover the tractable parts of the sim-to-real gap, but calibration drift and ALS--RGB spatial mismatch are unaddressed. The 117 D40 RAW captures (Sec.~4.6, Fig.~\ref{fig:realdata}) sanity-check the WP-only fallback only and the dataset is $84\%$ near-Planckian, where the ALS adds no information the white point does not already provide; end-to-end synchronized RGB$+$ALS validation on off-Planckian sources is the cleanest hardware-side next step. The single-illuminant SPD assumption also breaks down under indoor mixed lighting (e.g., daylight through a window plus an LED downlight). We test one camera (Nikon~D40); cross-sensor generalization is unverified, although principled dual-mapping schemes~\cite{yue2024dualmapping} suggest a tractable extension. Embedded-ISP latency on the per-shot NNLS$+$LSQ pipeline is not measured here.

%%%%%%%%%%%%%%%%%%%%%%%%%%%%%%%%%%
\section{Conclusion}
%%%%%%%%%%%%%%%%%%%%%%%%%%%%%%%%%%

Two-CCM interpolation by CCT is a 1D model of a higher-dimensional quantity. A few extra spectral observations recover SPD shape and fit a scene-specific CCM that approaches the per-illuminant Oracle within $0.3^{\circ}$ on Planckian, mild, and LED-like groups. Eight ALS channels suffice (a $K{=}8$ knee robust across basis sizes $M\!\in\![10,100]$). On the LED-like group ($n{=}45$) where the metameric problem is tangible, WP$+$ALS reaches a worst-25\% $1.7{\times}$ lower than the strongest non-spectral baseline (CST-MLP$_{3\times6}$); the lead is robust to AS7341-like filter shape and $5\%$ observation noise. Cross-database generalization requires augmenting the basis with class-representative reference SPDs; on Beyond RGB measured illuminants, the spectral lead is largest off the Planckian locus, as the metameric-problem framing predicts.

%%%%%%%%%%%%%%%%%%%%%%%%%%%%%%%%%%
% Bibliography
%%%%%%%%%%%%%%%%%%%%%%%%%%%%%%%%%%

\end{document}